%% file: main.tex
\PassOptionsToPackage{dvipsnames}{xcolor}

\documentclass[sigplan,nonacm,natbib=false]{acmart}
\usepackage{setspace}

\usepackage[
    backend=biber, 
    maxnames=5, 
    sorting=none,
    style=acmnumeric,
    maxcitenames=2,
]{biblatex}
\newcommand{\citet}[1]{\textcite{#1}}
\newcommand{\citep}[1]{\parencite{#1}}

\setcopyright{acmlicensed}
\copyrightyear{2018}
\acmYear{2018}
\acmDOI{XXXXXXX.XXXXXXX}

\acmConference[Conference acronym 'XX]{Make sure to enter the correct
  conference title from your rights confirmation emai}{June 03--05,
  2018}{Woodstock, NY}
\acmISBN{978-1-4503-XXXX-X/18/06}

\usepackage{acm_macros}

\hypersetup{colorlinks=true, citecolor=NavyBlue, linkcolor=NavyBlue, urlcolor=NavyBlue}

\begin{document}

\title{Invariant Pretraining for Robust Code Representations}

\author{Yifeng He}
\orcid{0000-0002-5389-7128}
\affiliation{%
	\institution{University of California at Davis}
	\city{Davis}
	\country{USA}
}
\email{yfhe@ucdavis.edu}

\author{Yundi Xu}
\affiliation{%
	\institution{University of California at Davis}
	\city{Davis}
	\country{USA}
}
\author{Christopher Castro Gaw Gonzalo}
\affiliation{%
	\institution{University of California at Davis}
	\city{Davis}
	\country{USA}
}
\author{Zili Wang}
\affiliation{%
	\institution{University of California at Davis}
	\city{Davis}
	\country{USA}
}

\author{Hao Chen}
\orcid{0000-0002-4072-0710}
\affiliation{%
	\institution{The University of Hong Kong}
	\country{China}
}
\email{chenho@hku.hk}


\input{src/abstract.tex}

\ccsdesc[500]{Security and privacy~Software security engineering}
\ccsdesc[500]{Computing methodologies~Neural networks}
\ccsdesc[300]{Computing methodologies~Unsupervised learning}
\keywords{Code Representation Learning, Model Robustness}

\maketitle

\input{src/introduction.tex}

\input{src/approach.tex}
\input{src/experiments.tex}
\input{src/related_work.tex}
\input{src/limitations.tex}
\input{src/conclusion.tex}

\section*{Acknowledgments}
We would like to thank the anonymous reviewers for their
constructive comments.
This work is supported by the UC Noyce Initiative.

\printbibliography

\end{document}

%% file: src/abstract.tex
\begin{abstract}
	Encoder-based code representation models remain widely deployed for discriminative tasks such as clone detection and code classification, where their small size and low inference cost are decisive.
	Their robustness, however, is fragile.
	On \emph{invariant programs}---semantically equivalent code written in different syntactic forms---their representations degrade substantially even though program behavior is unchanged.
	We present an empirical study of this robustness gap across four encoder baselines, two downstream tasks, and four datasets, together with a minimal code-only continued pretraining recipe that recovers a consistent part of it.
	Our method, invariant pretraining (\invpt), applies \spts to the corpus and combines masked language modeling with multi-positive supervised contrastive learning that treats all augmentations of the same source function as positives, mixing self-contrast pairs (same code, different masks) with invariant-contrast pairs (transformed code) for positives of varying difficulty.
	Unlike prior contrastive code encoders, \invpt does not require paired natural-language data.
	Across our evaluation, \invpt improves robustness on transformed test sets in every model--dataset comparison, by a median of $8.1$ percentage points on clone detection (up to $11.0$) and $3.6$ on code classification (up to $19.2$), while matching or improving standard accuracy; ablations isolate multi-positive invariant contrast as the main source of the gains.
	Because the transformed test sets compose the same operator family used in pretraining, we claim invariance to that family rather than robustness in general.
	Our aim is not a new objective but a careful measurement of where encoder robustness breaks and how far a simple, code-only recipe can recover it.
\end{abstract}

%% file: src/introduction.tex
\section{Introduction}

Machine learning for code is now dominated by large generative models~\citep{chen2021evaluating,rozière2023codellama,he2024unitsyn,lyu2024PromptFuzz}.
Encoder-based code representation models nonetheless remain the practical choice for discriminative tasks such as clone detection and code classification, which still sit inside modern pipelines: deduplicating training corpora, retrieving and ranking code, flagging near-duplicates in agent workflows.
Models such as CodeBERT~\citep{feng-etal-2020-codebert} and GraphCodeBERT~\citep{guo2021graphcodebert} are far smaller than generation models while performing well on these tasks.
With roughly 125M parameters,
they can match or outperform generation models in the 7B--15B range while requiring less storage and lower inference cost~\citep{thennal2025overparameterized,lei2025efficient}.
Their low fine-tuning cost keeps them attractive wherever robust code representations are needed at scale, so their failure modes still matter.

That robustness is not assured.
Modern programming languages are syntactically flexible:
developers routinely express the same computation in different ways because of coding style, conventions, or language idioms.
We refer to such semantically equivalent but syntactically varied programs as \emph{invariant programs} (or simply \emph{invariants}),
because their observable behavior is unchanged under these rewritings.
For example, a counting \texttt{for} loop can often be rewritten as an equivalent \texttt{while} loop without changing program behavior.
These variations are common in practice, yet code representation models often mishandle them because pretraining encourages sensitivity to surface tokens~\citep{yang2022natural_attack}.
\citet{Ahmed2022multilingual} showed that perturbing natural-language elements in code can substantially degrade model performance.
How large this degradation is across current encoders, and how much of it a simple training change can remove, has not been measured systematically.

Prior work addresses parts of the problem but does not settle it.
\citet{yu2022data} introduced SPAT, a \spt-based data-augmentation method that improved downstream performance without explicitly targeting robustness.
\citet{wang2022bridging} combined SPAT with curriculum learning, but focused on fine-tuning rather than pretraining.
\citet{liu2023contrabert} proposed contrastive pretraining, but relied on transformations that can alter semantics or introduce syntax errors, and on \emph{bi-modal} corpora with paired natural language and code (NL-PL pairs).
From CodeBERT~\citep{feng-etal-2020-codebert} to ContraBERT~\citep{liu2023contrabert}, these methods train on function--docstring pairs, and ContraBERT further designs natural-language transformations for its contrastive objective.
For robustness against invariant programs, however, paired NL-PL data is not essential:
the supervisory signal already lives in code through \spts.

\begin{figure*}[ht]
	\centering
	\includegraphics[width=\linewidth]{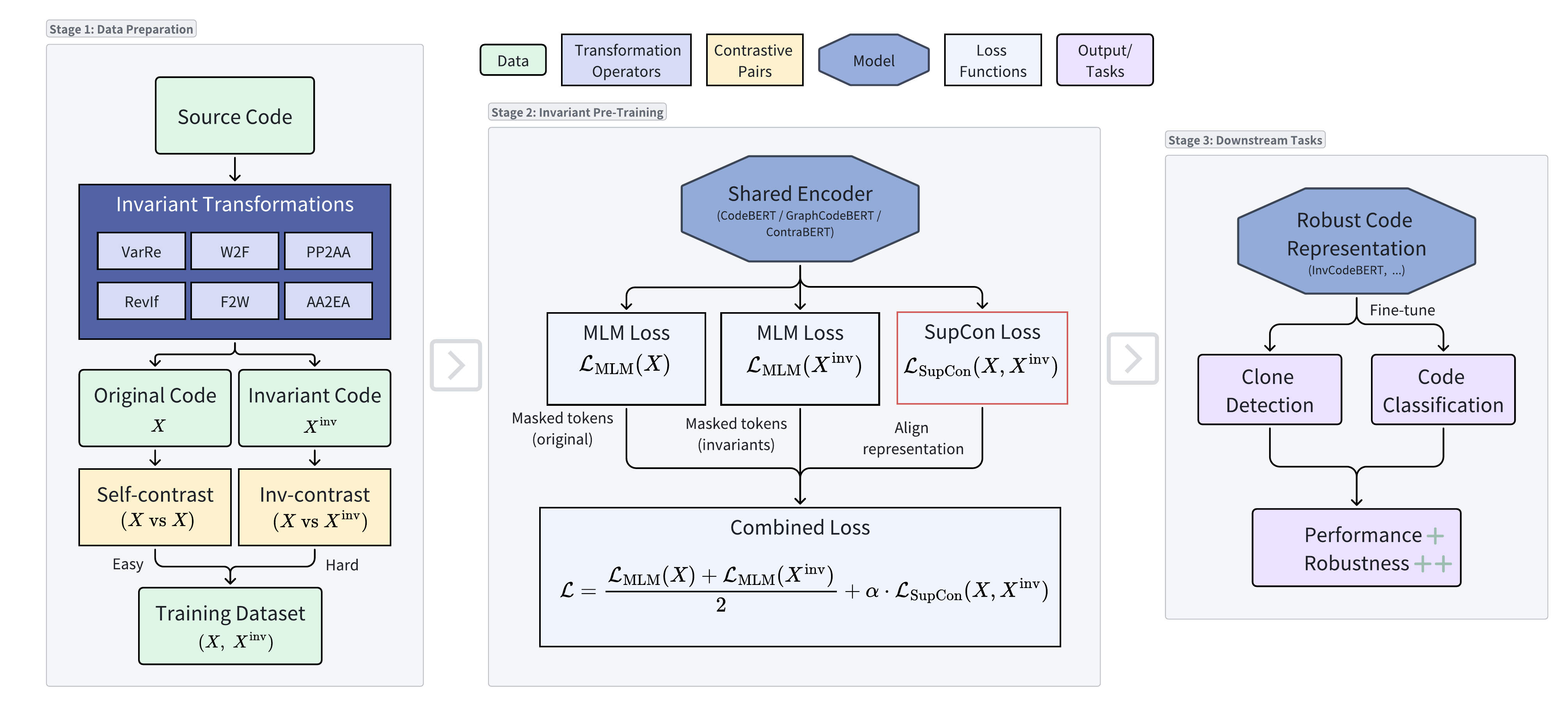}
	\caption{Overview of invariant pretraining (\invpt).}
	\label{fig:invpt}
\end{figure*}

We contribute an empirical study, not a new objective.
We quantify the robustness gap of encoder code models under invariant programs and ask how far a deliberately minimal, code-only recipe can close it.
The recipe, invariant pretraining (\invpt), is continued pretraining that uses only programming-language data.
\invpt applies \spts to the pretraining corpus (\autoref{sec:transformations}),
optimizes encoders with supervised contrastive learning (\autoref{sec:contrastive}),
and mixes self-contrast with invariant-contrast pairs to provide positives of varying difficulty (\autoref{sec:curriculum}).
Its only non-obvious design choice is a multi-positive contrastive mask that treats all augmentations of the same source function as positives,
avoiding the false-negative problem of standard InfoNCE,
which would push semantically equivalent variants apart.
This code-only design removes the need for paired NL-PL corpora and enables continued pretraining on large PL-only code collections.
The resulting models gain robustness on code-to-code tasks while preserving or improving standard downstream performance.

In summary, our contributions are:
\begin{enumerate}[label={\bf (\roman*)}]
	\item We measure the robustness of four encoder code models to \spts across clone detection and code classification on POJ104 and three CodeNet subsets (Java250, Python800, C++1400), quantifying a consistent degradation under semantically equivalent rewrites.
	\item We instantiate \invpt, a code-only continued pretraining recipe that combines invariant transformations with a multi-positive supervised contrastive objective.
	      It gains up to $6.57$ points of standard accuracy on the original clone-detection benchmark and improves robustness on every transformed model--dataset comparison, by a median of $8.07$ points on clone detection and $3.56$ on classification, recovering part but not most of the degradation.
	\item We run ablations isolating the source of improvement, showing that multi-positive invariant contrastive learning is the dominant factor, that self-contrast adds consistent gains, and that paired natural-language descriptions are unnecessary for invariant pretraining.
\end{enumerate}

%% file: src/approach.tex
\section{Invariant Pretraining}\label{sec:design}

\autoref{fig:invpt} overviews \invpt's three components: invariant code transformations, code-only contrastive learning, and mixed-difficulty contrastive pairs.
For each source snippet, we generate semantically equivalent but syntactically varied variants, and continue pretraining on both the original and its variants with a masked language modeling objective combined with an invariant contrastive objective.
Self-contrast and invariant-contrast examples are mixed so the model sees positives of varying difficulty during training.

\subsection{Invariant Code Transformations}\label{sec:transformations}

Invariant programs---semantically equivalent code that differs syntactically---arise
naturally from coding styles, conventions, and language idioms~\citep{ogura2018bring},
yet models often mishandle them because pretraining memorizes surface form~\citep{wang2022bridging,liu2023contrabert,carlini2019secret}.
We address this by pretraining with invariant code transformations
that produce semantically equivalent but syntactically diverse variants.
Following previous work on naturally occurring invariants~\citep{dui2021selfsupervised,yu2022data},
we design six transformation operators in two categories:
\emph{syntax} operators that modify surface-level tokens
and a \emph{branching} operator that alters control flow while preserving runtime behavior.
\autoref{tab:operators} lists each operator with its transformation rule and supported languages,
where $p$ denotes a predicate (condition) and $B$ a code block.
We describe each operator below with its preconditions and a correctness argument.

\begin{table*}[ht]
	\centering
	\caption{Invariant transformation operators, transformation rules, and supported languages.}
	\label{tab:operators}
	\begin{adjustbox}{max width=\textwidth}
		\input{tables/operators.tex}
	\end{adjustbox}
\end{table*}

We implement all operators at the AST level to ensure syntactic correctness:
we reuse the Java operators of \citet{yu2022data},
use \texttt{clang} for C/C++,
and the Python~3 \texttt{ast} module for Python
(with \texttt{2to3} as a fallback for Python~2 code~\citep{zhao2023understanding}).
We discard code snippets with syntax errors that prevent AST parsing.
Not all operators apply to every language (\eg,
Python's iterator-based for-loops preclude \wf and \fw),
as shown in \autoref{tab:operators}.
Our operators target the language constructs
that appear in our benchmarks (POJ104 and CodeNet);
for Python we therefore omit the loop and increment operators.

\paragraph{Variable Rename (\varre)}
Programmers choose variable names freely without affecting runtime behavior; a name may be a whole word, an abbreviation, a single character, a term in another language, or a random string.
\varre replaces all variable names with randomly generated, non-repeating strings that comply with the naming conventions of the target language, keeping the transformed code syntactically valid.
Variable names are purely syntactic identifiers with no effect on runtime behavior in any supported language (Java, C/C++, Python), and because \varre performs a consistent, injective renaming of all occurrences within each scope, data and control flow are unchanged.
Naming variation is pervasive in real code: style guides differ, internationalization yields non-English identifiers, and abbreviation conventions vary by domain~\citep{ogura2018bring}, so \varre models a naturally occurring transformation rather than an artificial adversarial perturbation.

\paragraph{While-to-For (\wf)}
While- and for-loops are both common imperative constructs that programmers use interchangeably.
Converting a while-loop to a for-loop reuses the condition while leaving the initialization and increment slots empty:
\[
	\texttt{while}\ (p)\ \{B\} \;\Rightarrow\; \texttt{for}\ (;\,p;\,)\ \{B\}.
\]
The for-loop \texttt{for(; p;)\{B\}} evaluates $p$ before each iteration and executes $B$ if true, identical to \texttt{while(p)\{B\}}; the empty slots add no computation, so semantics are preserved exactly.

\paragraph{For-to-While (\fw)}
The reverse converts a for-loop into a while-loop, which is more involved because while-loops lack a built-in counter.
We hoist the counter initialization before the loop and append the increment at the end of the body:
\[
	\texttt{for}\ (\mathit{init};\,p;\,\mathit{inc})\ \{B\} \;\Rightarrow\;
	\mathit{init};\ \texttt{while}\ (p)\ \{B;\,\mathit{inc}\}.
\]
\fw applies only to for-loops whose body has no \texttt{continue} statement: with \texttt{continue}, the appended $\mathit{inc}$ would be skipped, whereas the original for-loop runs $\mathit{inc}$ before the next condition check regardless.
Our \texttt{clang}-based implementation for C/C++ and the Java implementation of \citet{yu2022data} both check this precondition.
When $\mathit{init}$ introduces a declaration, we wrap the rewrite in a block (\ie, $\{\,\mathit{init};\ \texttt{while}\ (p)\ \{B;\,\mathit{inc}\}\,\}$) so loop-local declarations do not leak into the enclosing scope.
Under this precondition, $\mathit{init}$ runs once, $p$ is checked before each iteration, $B$ is the body, and $\mathit{inc}$ runs after each complete iteration, matching the for-loop semantics exactly.

\paragraph{PlusPlus-to-AddAssign (\ppaa)}
Self-increment operators are common and often interchangeable.
\ppaa converts an increment \texttt{x++} or \texttt{++x} into \texttt{x += 1} (in this direction only).
For post-increment we check on the parent AST node that the result value is not consumed, since \texttt{x++} evaluates to the old value whereas \texttt{x += 1} evaluates to the new one; pre-increment needs no such precondition.
We further restrict the operator to variables of primitive numeric type, so overloaded \texttt{operator++} in C++ is never rewritten.

\paragraph{AddAssign-to-EqualAdd (\aaea)}
\aaea converts ``add-assign'' into ``equal-add'': \texttt{x += 1} $\Rightarrow$ \texttt{x = x + 1}.
For primitive numeric types, which dominate our benchmarks (POJ104, CodeNet), \texttt{x += expr} and \texttt{x = x + expr} are equivalent in all supported languages.
For C/C++ the \texttt{clang}-based implementation enforces this precondition directly, rewriting only variables whose declared type is primitive numeric.
Python admits no such static check: the declared type is unavailable at the AST level, so our implementation rewrites every augmented assignment, and for objects that define \texttt{\_\_iadd\_\_}, such as lists and \texttt{numpy} arrays, \texttt{x += e} mutates in place while \texttt{x = x + e} rebinds a fresh object, which is observable under aliasing.
The rewrite is sound for the primitive-scalar arithmetic that dominates our competitive-programming corpora, but it is a heuristic rather than a guaranteed invariant on arbitrary Python (\autoref{sec:limitations}).

\paragraph{Reverse If (\revif)}
\revif reverses the condition of an \texttt{if} statement.
For a single-branch \texttt{if} (no \texttt{else}), we negate the condition, leave an empty branch, and move the original body to a new \texttt{else}:
\[
	\texttt{if}\ (p)\ \{B\} \;\Rightarrow\; \texttt{if}\ (\neg p)\ \{\}\ \texttt{else}\ \{B\}.
\]
For a two-branch \texttt{if}, we negate the condition and swap the branches:
\[
	\texttt{if}\ (p)\ \{B_1\}\ \texttt{else}\ \{B_2\} \;\Rightarrow\;
	\texttt{if}\ (\neg p)\ \{B_2\}\ \texttt{else}\ \{B_1\}.
\]
For \texttt{else if} chains (syntactic sugar for nested \texttt{if}s), we apply these rules recursively to each nested \texttt{if} node in the AST.
Negating $p$ and swapping branches preserves the mapping from condition values to blocks; for single-branch \texttt{if}s, the empty block has no effect; and recursive application is correct because each nested \texttt{if} is reversed independently.

\subsection{Invariant Contrastive Learning}\label{sec:contrastive}

Because our \spts preserve semantics under the preconditions of \autoref{sec:transformations}, the positive-pair gradient signal is unambiguous, and we use a single shared encoder rather than the momentum-decoupled encoder adopted by ContraBERT~\citep{liu2023contrabert,he2020momentum}.
We train with the average MLM loss on the original code and its invariant, combined with an invariant contrastive loss.
Since the transformations can produce multiple positives per anchor, we use supervised contrastive learning, which accommodates all positives without pushing equivalent augmentations apart as InfoNCE does.

\subsubsection{Masked Language Modeling}
Masked language modeling (MLM), a common pretraining objective for language models,
trains the model to predict masked tokens in the input sequence~\citep{devlin2019bert}.
Given an input sequence $X = \{[CLS], C, [EOS]\}$, where $C$ is the tokenized code snippet,
we randomly mask 15\% of the tokens in $X$ to obtain a masked sequence $X'$, following previous work~\citep{guo2021graphcodebert,liu2023contrabert,devlin2019bert,huang2024code}.
Let $M_X \subset X$ denote the masked tokens in $X$.
The loss is the negative log-likelihood of the masked tokens given $X'$:
\[
	\mlm(X) = - \frac{1}{|M_X|} \sum_{x \in M_X} \log P(x | X').
\]

\subsubsection{Supervised Contrastive Mask}
Because our dataset applies multiple transformation operators to the same source function (\eg, both \varre and \revif), different augmentations of the same function can co-occur in a mini-batch.
Under InfoNCE~\citep{oord2019representation}, these semantically equivalent augmentations would be \emph{pushed apart} as negatives, contradicting the invariance objective.
We therefore adopt supervised contrastive learning (SupCon)~\citep{khosla2020supervised} with a multi-positive mask: each snippet receives a deterministic identifier computed from its original source code (a truncated SHA-256 hash), so all augmentations of the same function share an identifier and are recognized as positives.

For a mini-batch of $B$ code--invariant pairs, we concatenate the $\ell_2$-normalized CLS embeddings into a pool of $N = 2B$ representations $\mathbf{z}$ with duplicated identifiers $\mathrm{ids}$, and form a multi-positive mask $M_{ij} = \mathbb{1}[\mathrm{ids}_i = \mathrm{ids}_j \wedge i \neq j]$.
Let $\mathcal{P}(i) = \{j : M_{ij} = 1\}$; every anchor has at least its paired augmentation as a positive, and when the same source function appears $k$ times each anchor has up to $2k-1$ positives.
We compute the identifier from the canonical, non-augmented source $c$ as
\[
	\mathrm{id}(c) = \mathrm{int64}\bigl(\mathrm{SHA\text{-}256}(c)[{:}8]\bigr)
	\mathbin{\&}\; \mathtt{0x7FFFFFFFFFFFFFFF},
\]
the first 8 bytes of the digest read as a big-endian 64-bit integer with the sign bit masked.
Because we hash the canonical source, all augmentations of a function, including self-contrast copies (\autoref{sec:curriculum}), share an identifier.
The assignment is deterministic and needs no coordination across training processes or data shards.
\autoref{tab:mask-example} illustrates the resulting mask for $B = 4$ pairs drawn from three source functions $A$, $B$, $C$, where $A$ appears twice through different operators (indices $0$--$3$ are original codes, $4$--$7$ their augmentations): anchors for $A$ (indices $0, 3, 4, 7$) each have $3$ positives, whereas anchors for $B$ and $C$ each have exactly $1$.

\begin{table}[t]
	\centering
	\caption{%
		Example positive mask $M$ for $B = 4$ with source functions $[A, B, C, A]$.
		Rows/columns 0--3 are original codes; 4--7 are their augmentations.
		``$\bullet$'': the positives ($M_{ij} = 1$);
		``$\text{-}$'': the excluded diagonal.
	}
	\label{tab:mask-example}
	\input{tables/mask-example.tex}
\end{table}

Let $\tau$ denote the temperature,
and let $\mathrm{sim}(\mathbf{u}, \mathbf{v}) = \mathbf{u}^{\mathsf{T}} \mathbf{v} / \tau$
be the scaled cosine similarity of $\ell_2$-normalized vectors.
The supervised contrastive loss is
\[
	\begin{aligned}
		\supcon =  - \frac{1}{|\mathcal{A}|} & \sum_{i \in \mathcal{A}}  \frac{1}{|\mathcal{P}(i)|} \\
		           & \sum_{p \in \mathcal{P}(i)}
		\log
		\frac{
			\exp\bigl(\mathrm{sim}(\mathbf{z}_i, \mathbf{z}_p)\bigr)
		}{
			\sum_{j \neq i}
			\exp\bigl(\mathrm{sim}(\mathbf{z}_i, \mathbf{z}_j)\bigr)
		},
	\end{aligned}
\]
where $\mathcal{A} = \{i : |\mathcal{P}(i)| > 0\}$ is the set of anchors with at least one positive.
When no same-function collisions occur, the mask reduces to the diagonal pairing of InfoNCE.
For a batch of code $X$ and invariants $\Xinv$, the overall objective is
\[
	\mathcal{L}(X, \Xinv) = \frac{\mlm(X) + \mlm(\Xinv)}{2} + \alpha \supcon,
\]
where $\alpha$ weights the contrastive loss.

\subsubsection{Mixed-Difficulty Contrastive Pairs}\label{sec:curriculum}

\paragraph{Self-contrast}
SimCSE~\citep{gao2021simcse} showed that passing the same input through an encoder with different dropout masks
produces effective contrastive pairs for sentence representation learning.
We introduce \emph{self-contrast} to code representation learning,
a signal that prior work~\citep{yu2022data,wang2022bridging,liu2023contrabert} has overlooked,
perhaps because small code edits can change semantics.
In \invpt, we include identical code snippets as contrastive pairs during pretraining,
relying on the $15\%$ MLM masking probability to produce varied positive pairs from the same code snippet.
Self-contrast provides the easier signal in our mixed-difficulty strategy:
the model learns to produce similar embeddings for the same code with different masked tokens.

\begin{table*}[h]
	\centering
	\caption{Correspondence between baseline models and our \invpt models.}
	\label{tab:model-names}
	\input{tables/model-names.tex}
\end{table*}

\paragraph{Invariant-contrast}
The single-operator transformations of \autoref{sec:transformations} supply the hard end of the range.
Whereas a self-contrast pair differs only in masked positions, an invariant pair can differ substantially in surface form while preserving semantics.
For example, \fw rewrites a \texttt{for}-loop into a \texttt{while}-loop by hoisting the initialization before the loop and appending the increment to the body, and \varre replaces every identifier with a fresh random name, so an anchor and its positive may share almost no tokens.
Aligning such a pair forces the model past surface cues toward the underlying computation, a strictly harder objective than tolerating masked tokens on an otherwise identical sequence.

\paragraph{Mixing difficulties}
We expose the model to both pair types simultaneously rather than staging them, presenting each anchor with its easy (self-contrast) and hard (invariant) positives in the same batch.
A staged schedule that withheld the harder pairs early and introduced them later would shift the positive-pair distribution mid-training, risking instability and catastrophic forgetting~\citep{kirkpatrick2017overcoming} during continued pretraining of a large language model.
We further start training with a low learning rate of \num{2e-5} so the model adapts gradually to the contrastive objective, then control the rate with standard warm-up and scheduling~\citep{kalra2024why}.
The ablation in \autoref{sec:ablation} supports this mix: keeping only the hard invariant pairs and removing self-contrast lowers performance on every original benchmark, so the easy signal complements the hard one rather than duplicating it.

%% file: tables/operators.tex
	\begin{tabular}{llll}
		\toprule
		Operator & Category  & Transformation rule                                                                                                                 & Lang.                   \\
		\midrule
		\varre   & Syntax    & Replace all variable names with random valid identifiers                                                                            & \iJava\;\iCpp\;\iPython \\
		\wf      & Syntax    & $\texttt{while}\ (p)\ \{B\} \;\Rightarrow\; \texttt{for}\ (;\,p;\,)\ \{B\}$                                                         & \iJava\;\iCpp           \\
		\fw      & Syntax    & $\texttt{for}\ (\mathit{init};\,p;\,\mathit{inc})\ \{B\} \;\Rightarrow\; \mathit{init};\ \texttt{while}\ (p)\ \{B;\,\mathit{inc}\}$ & \iJava\;\iCpp           \\
		\ppaa    & Syntax    & \texttt{x++} $\Rightarrow$ \texttt{x += 1}                                                                                    & \iJava\;\iCpp           \\
		\aaea    & Syntax    & \texttt{x += 1} $\Rightarrow$ \texttt{x = x + 1}                                                                                    & \iJava\;\iCpp\;\iPython \\
		\revif   & Branching & $\texttt{if}\ (p)\ \{B_1\}\ \texttt{else}\ \{B_2\} \;\Rightarrow\; \texttt{if}\ (\neg p)\ \{B_2\}\ \texttt{else}\ \{B_1\}$          & \iJava\;\iCpp\;\iPython \\
		\bottomrule
	\end{tabular}

%% file: tables/mask-example.tex
\small
\begin{tabular}{c cccc cccc}
	\toprule
	        & 0          & 1          & 2          & 3          & 4          & 5          & 6          & 7          \\
	\midrule
	0 ($A$) & $\text{-}$ &            &            & $\bullet$  & $\bullet$  &            &            & $\bullet$  \\
	1 ($B$) &            & $\text{-}$ &            &            &            & $\bullet$  &            &            \\
	2 ($C$) &            &            & $\text{-}$ &            &            &            & $\bullet$  &            \\
	3 ($A$) & $\bullet$  &            &            & $\text{-}$ & $\bullet$  &            &            & $\bullet$  \\
	\midrule
	4 ($A$) & $\bullet$  &            &            & $\bullet$  & $\text{-}$ &            &            & $\bullet$  \\
	5 ($B$) &            & $\bullet$  &            &            &            & $\text{-}$ &            &            \\
	6 ($C$) &            &            & $\bullet$  &            &            &            & $\text{-}$ &            \\
	7 ($A$) & $\bullet$  &            &            & $\bullet$  & $\bullet$  &            &            & $\text{-}$ \\
	\bottomrule
\end{tabular}

%% file: tables/model-names.tex
\begin{tabular}{lcccc}
	\toprule
	Baseline & CodeBERT & \gcb    & \contrabc   & \contrabg   \\
	\lightmidrule
	+ \invpt & \invcb   & \invgcb & \invcontrac & \invcontrag \\
	\bottomrule
\end{tabular}

%% file: src/experiments.tex
\section{Experiments}

\subsection{Tasks, Models, and Setup}
We evaluate \invpt on two code-to-code downstream tasks,
where robustness to invariant transformations is particularly important.
\textbf{Clone detection} retrieves semantically equivalent programs from a candidate pool
using cosine similarity over encoder embeddings.
We evaluate on POJ104~\citep{mou2016convolutional} (104 problems, 500 C/C++ solutions each)
and three CodeNet~\citep{puri2021codenet} subsets (Java250, Python800, and C++1400)
following~\citet{zhao2023understanding}.
\textbf{Code classification} predicts the functional category (problem number) of a program;
we evaluate on the same POJ104 and CodeNet benchmarks as clone detection.
\footnote{%
	We do not use BigCloneBench~\citep{svajlenko2014towards},
	due to label quality concerns~\citep{krinke2022bigclonebenchharmful,krinke2025misuse} and heavy data contamination.
	We also do not use Devign~\citep{zhou2019devign} defect detection
	due to serious data quality and label accuracy concerns~\citep{ding2024primevul}.
}

We compare CodeBERT~\citep{feng-etal-2020-codebert}, \gcb~\citep{guo2021graphcodebert},
\contrabc, and \contrabg~\citep{liu2023contrabert}, together with their \invpt counterparts.%
\footnote{We additionally include CodeSage~\citep{zhang2024codesage} and ModernBERT~\citep{warner2025smarter} as reference points in the clone detection evaluation (\autoref{tab:clone}). Since these models use different architectures, we do not apply \invpt continued pretraining to them.}
We follow the standard data splits and fine-tuning settings used in prior work~\citep{lu2021codexglue,puri2021codenet,zhao2023understanding};
\autoref{tab:model-names} lists the correspondence between each baseline and its \invpt variant.
For clone detection, we report Mean Average Precision at R (\mapr)~\citep{musgrave2020metric},
and for code classification we report accuracy.
Relative improvements (\%) over the corresponding baseline appear in gray.

We pretrain on the Java and Python subset of CodeSearchNet~\citep{husain2020codesearchnet}:
it provides high-coverage corpora for both languages, and all six transformation operators apply to them.
We hold C/C++ out of pretraining on purpose.
The cross-language evaluation in \autoref{sec:rq1} and \autoref{sec:robustness} then tests whether structural invariance learned on Java and Python transfers to a disjoint target language, rather than measuring in-distribution performance.
We pretrain with maximum sequence length 512, batch size 256, temperature $\tau = 0.1$, and contrastive weight $\alpha = 1.0$,
using AdamW~\citep{loshchilov2019decoupledweightdecayregularization} with learning rate \num{2e-5}, weight decay \num{0.01}, and $10\%$ warmup,
for 3 epochs with checkpoints selected by validation loss.
For downstream fine-tuning, we follow CodeXGLUE~\citep{lu2021codexglue} for the POJ104 split and \citet{puri2021codenet} for the CodeNet splits,
using learning rate \num{2e-5}, batch size 8, and maximum sequence length 400.
All pretraining runs use four NVIDIA H100 GPUs with 80~GB memory each, and all downstream runs use a single NVIDIA H100 GPU.
Continued pretraining is inexpensive relative to pretraining from scratch: each \invpt model takes about 20 hours on the four H100s, or roughly 80 GPU-hours, for the three epochs over the augmented corpus.
All reported numbers come from a single pretraining and fine-tuning run per configuration, with no repeats across random seeds (\autoref{sec:limitations}).

We organize the evaluation around the claims made in the introduction.
Throughout, we use \emph{standard accuracy} to refer to accuracy on the original (untransformed) test set, in contrast to robustness, which we measure on the transformed test sets.
We first test whether \invpt preserves standard downstream performance.
We then evaluate robustness on transformed test sets to measure invariance under unseen compositions of transformations.
Next, we use ablations to identify which design components drive the gains.
Finally, we present representation visualizations as qualitative evidence that is consistent with the quantitative results.

\subsection{Standard Downstream Performance}\label{sec:rq1}

\begin{table*}[t]
	\centering
	\caption{Clone detection results.
		``orig'': original test set; ``+T'': transformed test set for robustness evaluation.}
	\label{tab:clone}
	\begin{adjustbox}{max width=\textwidth}
		\input{tables/clone_merged.tex}
	\end{adjustbox}
\end{table*}

We first evaluate standard downstream performance to verify that \invpt does not buy robustness at the cost of standard accuracy.
We then turn to transformed test sets in \autoref{sec:robustness},
where robustness to invariant programs is the main target.
\autoref{tab:clone} shows the clone detection results.
Applying \invpt improves all four baseline families on most benchmarks,
and \invgcb and \invcontrag achieve the strongest overall performance.
The largest gain appears on POJ104, where \invgcb improves \gcb by $7.8\%$.
Notably, these gains also transfer to the C/C++ benchmarks.
Although the base encoders were pretrained on C/C++ (among other languages),
our invariant continued pretraining uses \emph{only} Java and Python programs from CodeSearchNet;
no C/C++ code receives any invariant transformation during this stage.
The C/C++ improvements therefore reflect cross-language transfer of the robustness signal acquired from Java and Python invariants alone.

Code classification accuracy is nearly saturated on these benchmarks (orig columns of \autoref{tab:classification}).
Even in this regime, \invpt matches or slightly improves the corresponding baselines on all four datasets.

\begin{table*}[t]
	\centering
	\caption{Code classification results.
		``orig'': original test set; ``+T'': transformed test set for robustness evaluation.}
	\label{tab:classification}
	\begin{adjustbox}{max width=\textwidth}
		\input{tables/classification_merged.tex}
	\end{adjustbox}
\end{table*}

\subsection{Robustness Against Invariant Programs}\label{sec:robustness}

We evaluate robustness by cumulatively applying all six transformations from \autoref{sec:transformations}
to the test set.
Because training uses only single-operator transformations, these compositions are unseen, so the setting measures invariance to the pretraining rewrite family under novel compositions rather than robustness to arbitrary semantics-preserving edits (\autoref{sec:limitations}).
For clone detection, all \invpt models outperform their baselines on transformed data (\autoref{tab:clone}),
with relative gains up to $22.2\%$.
\invgcb also surpasses both \contrabc and \contrabg on all four benchmarks, and \invcb on Java250, Python800, and C++1400; the one exception is POJ104, where \invcb ($62.03$) remains below \contrabc ($62.76$) and \contrabg ($64.20$).

For code classification, \invpt again yields consistent gains on transformed data (+T columns of \autoref{tab:classification}),
improving over \gcb by up to $7.7\%$ and over \contrabg by up to $6.1\%$.
Notably, these gains extend to C++1400
(up to $+3.77$ on classification and $+5.84$ on clone detection),
where no C/C++ code receives invariant transformations during pretraining,
demonstrating \emph{cross-language transfer} of structural robustness from Java and Python to a held-out target language.
We do not claim cross-lingual representational alignment (\ie, mapping semantically equivalent programs across languages into nearby embeddings), which would require parallel multi-language data and is left to future work.

\paragraph{Where \invpt helps least}
The smallest robustness gain on the transformed test sets is \invcontrac vs.\ \contrabc on POJ104 ($+0.77$ pp, $+1.2\%$), indicating that \contrabc's prior contrastive objective already captures much of the invariance signal on this comparatively saturated benchmark.
The gain is largest where the baseline is weakest: on transformed C++1400, where C/C++ never appears in our pretraining corpus, \invcb improves over CodeBERT by $+5.84$ pp ($+22.8\%$).

\paragraph{Why cross-language transfer works}
We attribute the transfer to two factors.
First, our operators target language-agnostic structural concepts: loop equivalence (\wf, \fw), branch reversal (\revif), scalar increment forms (\ppaa, \aaea), and identifier renaming (\varre).
Second, Java, Python, and C/C++ share a lexical substrate (common keywords, operators, and bracketing) that the encoder's sub-word tokenizer represents consistently across languages.
We do not provide a theoretical analysis of this transfer; these two factors are the strongest explanation our evidence supports.

\paragraph{Absolute percentage-point improvements}
The tables above report performance and relative improvement (\%) over each baseline.
For readers who prefer absolute changes, \autoref{tab:clone-pp} and \autoref{tab:classification-pp} restate the robustness gains on the transformed (+T) test sets in percentage points (pp),
the simple arithmetic difference between each \invpt model and its corresponding baseline
(\eg, \invcb on Java250 (+T): a $+8.11$~pp gain over CodeBERT).
\invpt improves in all $16$ model--dataset comparisons on each task, but the distributions differ sharply.
On clone detection the gains are uniformly sizable (median $+8.07$~pp, up to $+11.04$).
On classification they are smaller and strongly skewed (median $+3.56$~pp), with the largest value ($+19.16$, \invcb on Python800) well separated from the next ($+10.67$); we therefore treat clone detection as the more representative measure and the classification maximum as an outlier rather than a typical gain.
Measured against the degradation itself, \invpt recovers only part of the drop from the original to the transformed test set: a median of $29.6\%$ on clone detection (range $2.9$--$49.3\%$) and $8.7\%$ on classification (range $1.4$--$53.7\%$), so the gap that remains is larger than the part we close.

\begin{table}[t]
	\centering
	\caption{Clone detection (+T): absolute pp change vs.\ the corresponding baseline encoder (\invcb vs.\ CodeBERT, \invgcb vs.\ \gcb, etc.).}
	\label{tab:clone-pp}
	\begin{adjustbox}{max width=\columnwidth}
		\input{tables/clone+T_pp.tex}
	\end{adjustbox}
\end{table}

\begin{table}[t]
	\centering
	\caption{Code classification (+T): absolute pp change vs.\ the corresponding baseline encoder.}
	\label{tab:classification-pp}
	\begin{adjustbox}{max width=\columnwidth}
		\input{tables/classification+T_pp.tex}
	\end{adjustbox}
\end{table}

\subsection{Ablation Study}\label{sec:ablation}

\begin{table*}[t]
	\centering
	\caption{Ablation studies on clone detection using CodeNet and POJ104.
		``orig'': original test set; ``+T'': transformed test set.} 
	\label{tab:ablation}
		\input{tables/ablations.tex}
\end{table*}

We ablate the contribution of each component of \invpt (\autoref{sec:design}) with three CodeBERT-based variants:
\begin{enumerate*}
	\item \textbf{w/o contrastive loss}: drops the supervised contrastive objective and trains with only MLM on original and transformed code;
	\item \textbf{w/o self-contrast}: keeps only invariant-contrast pairs;
	\item \textbf{\invcb $+$ NL}: adds natural-language descriptions alongside code in pretraining.
\end{enumerate*}
We evaluate the variants on clone detection (CodeNet, POJ104) with original and transformed (+T) test sets, and report the full result in \autoref{tab:ablation}.

\paragraph{Contrastive learning is the primary driver of improvement}
Removing the contrastive loss and training only with MLM on original and transformed code
still improves over CodeBERT on most benchmarks,
but it remains clearly below full \invcb.
On the original POJ104,
it even drops below the CodeBERT baseline ($83.87$ vs.\ $85.47$),
indicating that exposure to transformed code without explicit representation alignment is insufficient.

\paragraph{Self-contrast provides consistent additional gains}
Removing self-contrast lowers performance on all original benchmarks relative to \invcb,
by roughly $0.5$ to $1.1$ points,
and usually also reduces robustness on transformed test sets.
The margins are modest,
but the pattern is stable across datasets,
suggesting that aligning different masked views of the same program
complements invariant contrast rather than duplicating it.

\paragraph{Natural language descriptions are not necessary}
Adding natural language descriptions yields mixed results.
The effect is inconsistent across benchmarks and small relative to the other components: on the transformed sets, $+$NL gains $+2.08$~pp on both Java250 and POJ104 but loses $0.68$ and $0.42$~pp on Python800 and C++1400, averaging $0.76$~pp above \invcb on transformed data ($56.58$ vs.\ $55.82$) and $0.18$~pp below it on original data ($78.49$ vs.\ $78.67$).
We therefore do not claim that code-only pretraining outperforms $+$NL.
The ablation supports only a weaker claim: paired natural language moves performance by well under a point on average, and in opposite directions on the two settings.
That gain does not justify a bi-modal NL-PL corpus, which restricts continued pretraining to the comparatively small set of functions carrying usable docstrings.
\autoref{tab:ablation-pp} restates the ablation effects on the transformed (+T) test sets as absolute pp changes relative to full \invcb, making each removal's magnitude directly comparable: dropping the contrastive loss costs the most ($-2.28$ to $-4.52$~pp), while removing self-contrast or adding NL descriptions matters less.

\begin{table}[t]
	\centering
	\caption{Ablations (+T) absolute pp change.}
	\label{tab:ablation-pp}
	\begin{adjustbox}{max width=\columnwidth}
		\input{tables/ablations_pp.tex}
	\end{adjustbox}
\end{table}

\begin{figure*}[ht]
	\centering
	\includegraphics[width=\textwidth]{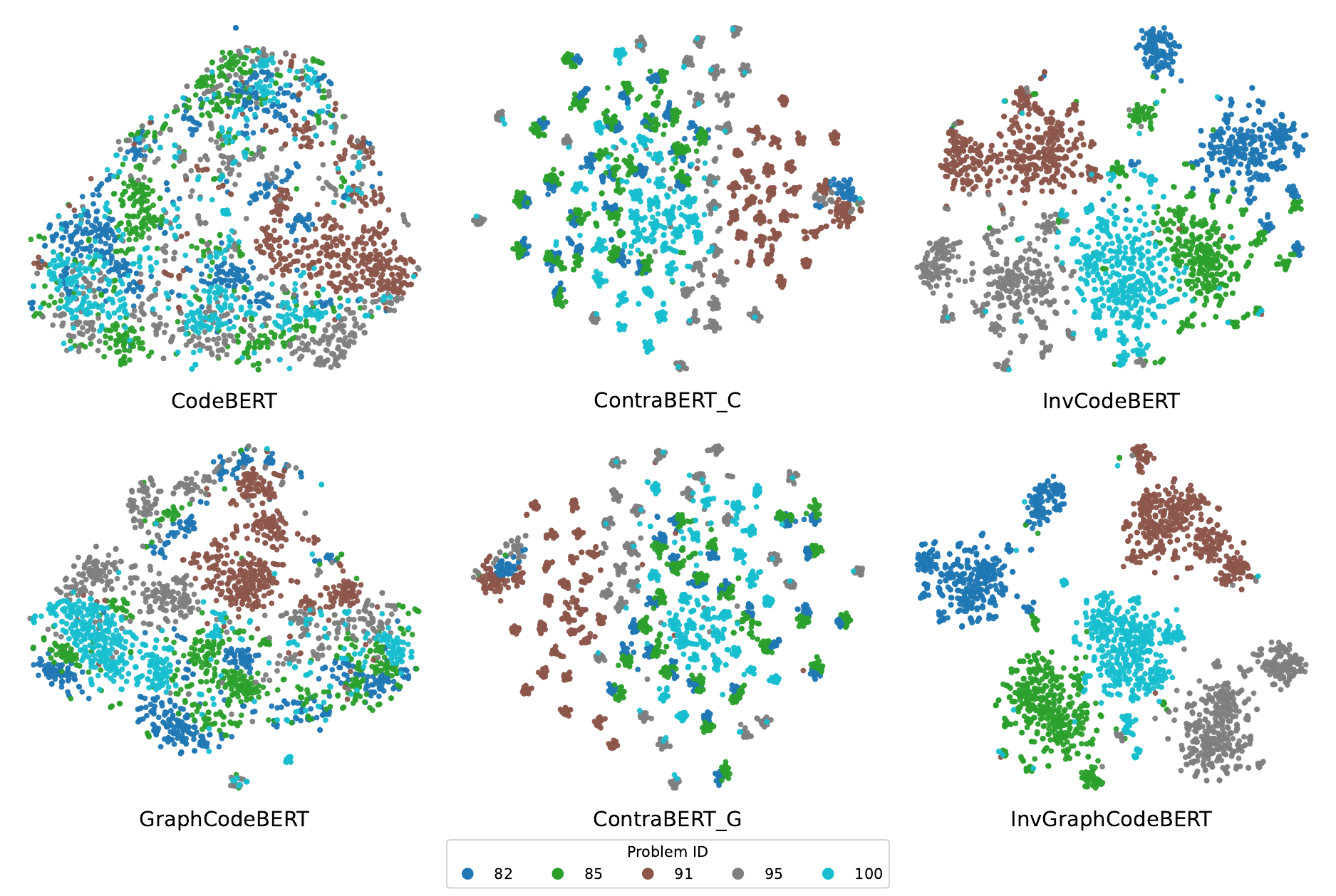}
	\caption{Visualization of vector embeddings of 5 problems in POJ104.
		Top row: CodeBERT-based models; bottom row: GraphCodeBERT-based models.}
	\label{fig:cluster}
\end{figure*}

\subsection{Qualitative Representation Visualization}
We next present a qualitative view of the learned representation space.
Following \citet{liu2023contrabert}, we randomly select five problems from the test split of POJ104,
apply invariant transformations to the code snippets within these problems,
and use all 500 code snippets per problem for visualization.
We then apply t-SNE~\citep{tSNE} to project the high-dimensional code representations into two dimensions.
\autoref{fig:cluster} shows the results for two model families.

Across both families, the base models (CodeBERT and \gcb)
produce heavily overlapping embeddings with little cluster separation:
invariant-transformed programs often lie closer to unrelated solutions than to their semantic equivalents.
The ContraBERT variants (\contrabc and \contrabg)
introduce more structure, with same-problem points forming local groupings,
though considerable overlap remains.
In contrast, our \invpt models (\invcb and \invgcb)
yield tight, well-separated clusters for each problem,
providing qualitative evidence consistent with the quantitative results. 

%% file: tables/clone_merged.tex
\begin{tabular}{l N N@{}l N N@{}l N N@{}l N N@{}l}
	\toprule
	                 & \multicolumn{3}{c}{Java250} & \multicolumn{3}{c}{Python800} & \multicolumn{3}{c}{C++1400} & \multicolumn{3}{c}{POJ104}                                                                                        \\
	\cmidrule(lr){2-4} \cmidrule(lr){5-7} \cmidrule(lr){8-10} \cmidrule(lr){11-13}
	Model            & {orig} & \multicolumn{2}{c}{+ T} & {orig} & \multicolumn{2}{c}{+ T} & {orig} & \multicolumn{2}{c}{+ T} & {orig} & \multicolumn{2}{c}{+ T}                                                                   \\
	\midrule
	CodeBERT         & 77.28  & 51.16              &              & 83.71              & 62.17              &              & 56.21              & 25.59              &              & 85.47              & 55.96              &              \\
	GraphCodeBERT    & 82.15  & 53.52              &              & 86.06              & 65.89              &              & 58.76              & 30.26              &              & 84.09              & 58.63              &              \\
	\contrabc        & 82.19  & 49.64              &              & 85.60              & 63.41              &              & 57.65              & 27.09              &              & 89.58              & 62.76              &              \\
	\contrabg        & 83.47  & 53.79              &              & 85.97              & 65.98              &              & 59.15              & 29.11              &              & 90.80              & 64.20              &              \\
	CodeSage         & 72.43  & 25.92              &              & 61.11              & 25.89              &              & 51.53              & 21.33              &              & 78.20              & 52.87              &              \\
	ModernBERT       & 80.51  & 41.06              &              & 83.16              & 52.85              &              & 59.05              & 24.66              &              & 88.22              & 52.86              &              \\
	\lightmidrule
	InvCodeBERT      & 82.44  & 59.27              & \rimp{+15.9} & 85.63              & 70.55              & \rimp{+13.5} & 57.80              & 31.43              & \rimp{+22.8} & 88.81              & 62.03              & \rimp{+10.8} \\
	InvGraphCodeBERT & 84.41  & \bfseries 62.30    & \rimp{+16.4} & \bfseries 86.73    & 73.91              & \rimp{+12.2} & \bfseries 59.23    & \bfseries 33.68    & \rimp{+11.3} & 90.66              & 67.32              & \rimp{+14.8} \\
	InvContraBERT\_C & 82.63  & 60.68              & \rimp{+22.2} & 86.16              & 74.34              & \rimp{+17.2} & 58.09              & 32.07              & \rimp{+18.4} & 89.28              & 63.53              & \rimp{+1.2}  \\
	InvContraBERT\_G & \bfseries 84.53 & 62.24     & \rimp{+15.7} & 86.34              & \bfseries 74.60    & \rimp{+13.1} & 59.23              & 33.10              & \rimp{+13.7} & \bfseries 91.25    & \bfseries 68.63    & \rimp{+6.9}  \\
	\bottomrule
\end{tabular}

%% file: tables/classification_merged.tex
\begin{tabular}{l N N@{}l N N@{}l N N@{}l N N@{}l}
	\toprule
	                 & \multicolumn{3}{c}{Java250} & \multicolumn{3}{c}{Python800} & \multicolumn{3}{c}{C++1400} & \multicolumn{3}{c}{POJ104}                                                                                        \\
	\cmidrule(lr){2-4} \cmidrule(lr){5-7} \cmidrule(lr){8-10} \cmidrule(lr){11-13}
	Model            & {orig} & \multicolumn{2}{c}{+ T} & {orig} & \multicolumn{2}{c}{+ T} & {orig} & \multicolumn{2}{c}{+ T} & {orig} & \multicolumn{2}{c}{+ T}                                                                   \\
	\midrule
	CodeBERT         & 97.80           & 65.81           &              & 98.92           & 63.26           &              & 89.63           & 39.92           &             & 98.51           & 50.97           &             \\
	GraphCodeBERT    & 97.97           & 67.51           &              & 99.06           & 78.52           &              & 89.70           & 41.52           &             & 98.68           & 52.03           &             \\
	\contrabc        & 97.90           & 67.33           &              & 98.89           & 72.61           &              & 89.54           & 41.46           &             & 98.48           & 51.19           &             \\
	\contrabg        & 98.15           & 68.57           &              & 99.10           & 78.59           &              & 89.53           & 41.23           &             & 98.63           & 50.38           &             \\
	\lightmidrule
	InvCodeBERT      & 98.16           & 70.10           & \rimp{+6.5}  & 99.00           & 82.42           & \rimp{+30.3} & 89.94           & 43.69           & \rimp{+9.4} & 98.63           & 53.67           & \rimp{+5.3} \\
	InvGraphCodeBERT & 98.08           & 71.30           & \rimp{+5.6}  & 99.09           & \bfseries 84.60 & \rimp{+7.7}  & \bfseries 90.00 & \bfseries 44.23 & \rimp{+6.5} & 98.54           & 52.66           & \rimp{+1.2} \\
	InvContraBERT\_C & 98.07           & 71.35           & \rimp{+6.0}  & 98.95           & 83.28           & \rimp{+14.7} & 89.87           & 42.51           & \rimp{+2.5} & 98.72           & 53.24           & \rimp{+4.0} \\
	InvContraBERT\_G & \bfseries 98.21 & \bfseries 71.48 & \rimp{+4.2}  & \bfseries 99.11 & 83.42           & \rimp{+6.1}  & 89.71           & 42.88           & \rimp{+4.0} & \bfseries 98.73 & \bfseries 53.74 & \rimp{+6.7} \\
	\bottomrule
\end{tabular}

%% file: tables/clone+T_pp.tex
\begin{tabular}{l r r r r}
	\toprule
	Model            & {Java250} & {Python800} & {C++1400} & {POJ104} \\
	\midrule
	InvCodeBERT      & $+8.11$   & $+8.38$     & $+5.84$   & $+6.07$  \\
	InvGraphCodeBERT & $+8.78$   & $+8.02$     & $+3.42$   & $+8.69$  \\
	InvContraBERT\_C & $+11.04$  & $+10.93$    & $+4.98$   & $+0.77$  \\
	InvContraBERT\_G & $+8.45$   & $+8.62$     & $+3.99$   & $+4.43$  \\
	\bottomrule
\end{tabular}

%% file: tables/classification+T_pp.tex
\begin{tabular}{l r r r r}
	\toprule
	Model            & {Java250} & {Python800} & {C++1400} & {POJ104} \\
	\midrule
	InvCodeBERT      & $+4.29$   & $+19.16$    & $+3.77$   & $+2.70$  \\
	InvGraphCodeBERT & $+3.79$   & $+6.08$     & $+2.71$   & $+0.63$  \\
	InvContraBERT\_C & $+4.02$   & $+10.67$    & $+1.05$   & $+2.05$  \\
	InvContraBERT\_G & $+2.91$   & $+4.83$     & $+1.65$   & $+3.36$  \\
	\bottomrule
\end{tabular}

%% file: tables/ablations.tex
\begin{tabular}{l NN NN NN NN}
	\toprule
	                     & \multicolumn{2}{c}{Java250} & \multicolumn{2}{c}{Python800} & \multicolumn{2}{c}{C++1400} & \multicolumn{2}{c}{POJ104}                                   \\
	\cmidrule(lr){2-3} \cmidrule(lr){4-5} \cmidrule(lr){6-7} \cmidrule(lr){8-9}
	Model                & {orig}                      & {+ T}                         & {orig}                      & {+ T}                      & {orig} & {+ T} & {orig} & {+ T} \\
	\midrule
	CodeBERT             & 77.28                       & 51.16                         & 83.71                       & 62.17                      & 56.21  & 25.59 & 85.47  & 55.96 \\
	\lightmidrule
	w/o contrastive loss & 80.16                       & 56.12                         & 84.55                       & 68.27                      & 56.05  & 28.47 & 83.87  & 57.51 \\
	w/o self-contrast    & 81.92                       & 57.92                         & 85.07                       & 70.27                      & 57.19  & 30.33 & 87.69  & 62.42 \\
	\invcb{} $+$ NL      & 82.74                       & 61.35                         & 85.52                       & 69.87                      & 57.69  & 31.01 & 88.02  & 64.11 \\
	\lightmidrule
	\invcb               & 82.44                       & 59.27                         & 85.63                       & 70.55                      & 57.80  & 31.43 & 88.81  & 62.03 \\
	\bottomrule
\end{tabular}

%% file: tables/ablations_pp.tex
\begin{tabular}{l r r r r}
	\toprule
	Variant              & {Java250} & {Python800} & {C++1400} & {POJ104} \\
	\midrule
	w/o contrastive loss & $-3.15$   & $-2.28$     & $-2.96$   & $-4.52$  \\
	w/o self-contrast    & $-1.35$   & $-0.28$     & $-1.10$   & $+0.39$  \\
	\invcb{} $+$ NL      & $+2.08$   & $-0.68$     & $-0.42$   & $+2.08$  \\
	\bottomrule
\end{tabular}

%% file: src/related_work.tex
\section{Related Work}

\paragraph{Pretrained code models}
Code representation learning adapts encoder pretraining to programming languages,
with models such as CodeBERT~\citep{feng-etal-2020-codebert} and \gcb~\citep{guo2021graphcodebert}
learning from large code corpora~\citep{he2024unitsyn,he2025fuzzaug}.
Subsequent work adds
AST sequences~\citep{guo2022unixcoder},
data-flow graphs~\cite{guo2021graphcodebert},
execution signals~\citep{huang2024code},
and transformation-based augmentation~\citep{yu2022data}.
None of them explicitly trains for invariant embeddings
under semantics-preserving transformations.
Recent work confirms that encoders remain competitive and efficient
for understanding tasks~\citep{thennal2025overparameterized,lin2026cl4d,warner2025smarter,lei2025efficient},
motivating \invpt's encoder-based design.
A complementary line of work scales code embedding models well beyond the 125M-parameter class~\citep{jina_code_embeddings_2025,qwen3_embedding_2025,embeddinggemma_2025,coderankembed_2024,wang2023codet5p};
\invpt is architecture-agnostic, and we instantiate it on 125M-parameter encoders that dominate cost-sensitive deployments while including CodeSage~\citep{zhang2024codesage} (1.3B) and ModernBERT~\citep{warner2025smarter} as larger references in \autoref{tab:clone}.

\paragraph{Contrastive learning for code}
Contrastive learning is a dominant paradigm
for transferable representations;
SimCLR~\citep{pmlr-v119-chen20j} established that augmentation composition
and a learnable projection head are critical ingredients.
In the code domain,
VarCLR~\citep{chen2022varclr} and CodeSage~\citep{zhang2024codesage}
show that contrasting transformed code improves representation quality,
while NatGen~\citep{chakraborty2022natgen} applies similar \spts
in a generative denoising setting.

Two lines of work share our core premise that semantics-preserving transformations supply a contrastive signal for code:
Corder~\citep{dui2021selfsupervised} contrasts AST-level source-to-source variants, and ContraCode~\citep{jain2021contrastive} contrasts compiler-generated JavaScript variants.
We do not claim that premise as a contribution.
\invpt differs in three respects.
Both prior methods use pairwise InfoNCE, which treats two variants of the \emph{same} function as negatives when they co-occur in a batch; since we apply several operators per function, such collisions are frequent, and we instead use a multi-positive supervised contrastive mask keyed on a hash of the canonical source (\autoref{sec:contrastive}).
We apply the recipe as \emph{continued} pretraining on released encoders rather than pretraining from scratch, and we evaluate robustness explicitly, on transformed test sets and a held-out target language, rather than standard accuracy alone.
We do not rerun their objectives on our backbone, so we report no controlled comparison; our ablation isolates the contrastive objective as a whole (\autoref{sec:ablation}) but not the multi-positive mask against pairwise InfoNCE.

The closest prior work is ContraBERT~\citep{liu2023contrabert}, which combines MoCo with bi-modal NL-PL pretraining but evaluates robustness primarily on variable renaming.
\invpt differs by using code-only continued pretraining with supervised contrastive learning over a shared encoder and a multi-positive mask, targeting naturally occurring \spts (loop equivalence, branch reversal, scalar-increment forms, variable renaming) rather than dead-code insertion and random line deletion, and evaluating across multiple transformation types and downstream tasks.
SPAT~\citep{yu2022data,wang2022bridging} uses similar \spts for fine-tuning augmentation; \invpt instead bakes them into pretraining as an objective. 

\paragraph{Adversarial robustness}
DAMP~\citep{yefet2020damp}, MHM~\citep{zhang2020generating}, and ALERT~\citep{yang2022natural_attack}
use variable renaming and dead-code insertion as adversarial attacks on code models;
\citet{bielik2020adversarial} give a systematic treatment of code adversarial robustness,
\citet{henkel2022semantic} formalize $k$-transformation robustness,
and \citet{rabin2021generalizability} frame it as generalizability under natural transformations.
Unlike these fine-tuning defenses, \invpt builds invariance into pretraining, amortizing the benefit across downstream tasks.

%% file: src/limitations.tex
\section{Limitations}\label{sec:limitations}

Our robustness evaluation uses test-time compositions that are unseen but built from the same six operators used in pretraining.
The results therefore establish invariance to this rewrite family, not robustness to semantics-preserving change in general; independently generated transformations and naturally occurring refactorings remain untested.
We also do not compare against downstream-only augmentation, so we cannot separate building invariance into pretraining from augmenting the fine-tuning data with the same rewrites.

All results come from a single pretraining and fine-tuning run per configuration, so differences of a point or less---including several ablation margins in \autoref{tab:ablation}---should be read with caution.

Our operators cover Java, Python, and C/C++; extending them to languages with substantially different semantics (\eg, Haskell, Rust) requires non-trivial engineering to preserve equivalence.
Because the rewrites are automated rather than formally verified, rare edge cases may alter runtime behavior, the clearest being \aaea on Python (\autoref{sec:transformations}); we expect the affected fraction of our corpora to be small but have not measured it.

Finally, we evaluate discriminative tasks on competitive-programming benchmarks, which provide ground-truth labels but under-represent industrial codebases, and we instantiate \invpt on 125M-parameter encoder-only models pretrained on CodeSearchNet.
Invariance is the appropriate criterion only for tasks whose output should not change under rewriting; generation requires the more general equivariance condition, which we leave to future work.

%% file: src/conclusion.tex
\section{Conclusion}

We present \invpt, a code-only continued pretraining method that combines \spts with supervised contrastive learning and mixed-difficulty positive pairs, eliminating the need for paired natural-language data.
\invpt improves robustness over its baseline in every model--dataset comparison on clone detection and code classification, by a median of $8.07$ and $3.56$ percentage points respectively (up to $11.04$ and $19.16$), while matching or improving standard task performance.
These gains recover part of the degradation, not most of it, and are measured on compositions of the operator family seen in pretraining; robustness to structurally different rewrites remains open.
Ablations show that invariant contrastive learning is the primary driver of these gains, with self-contrast providing consistent additional improvement.
Notably, pretraining on Java and Python invariants alone yields robustness gains on C/C++ downstream tasks,
suggesting that the learned invariance transfers across languages.
More broadly, \invpt demonstrates that robust code representations can be learned from code alone.
Because the method modifies only the pretraining data and loss, extending it to richer transformation families and to encoder-decoder or decoder-only architectures is a natural next step.
Generation, however, calls for a weaker criterion than invariance: under \varre a model should rename consistently in its output rather than leave it unchanged.
The general condition for a transformation $T$ and a model $g$ is equivariance, $g(T(x)) = M_T(g(x))$ for a transformation-specific output map $M_T$, with invariance the special case $M_T = \mathrm{id}$; learning equivariance under a known $M_T$ is the direction we see as most promising.

%% file: main.bib
@inproceedings{zhang2024codesage,
  title     = {Code Representation Learning at Scale},
  author    = {Zhang, Dejiao and Ahmad, Wasi and Tan, Ming and Ding, Hantian and Nallapati, Ramesh and Roth, Dan and Ma, Xiaofei and Xiang, Bing},
  booktitle = {International Conference on Learning Representations},
  editor    = {B. Kim and Y. Yue and S. Chaudhuri and K. Fragkiadaki and M. Khan and Y. Sun},
  pages     = {47427--47446},
  url       = {https://proceedings.iclr.cc/paper_files/paper/2024/file/cfbba5249393100ada0bfb37557d2fd9-Paper-Conference.pdf},
  volume    = {2024},
  year      = {2024}
}

@inproceedings{warner2025smarter,
  title     = {Smarter, Better, Faster, Longer: A Modern Bidirectional Encoder for Fast, Memory Efficient, and Long Context Finetuning and Inference},
  author    = {Warner, Benjamin  and
               Chaffin, Antoine  and
               Clavi{\'e}, Benjamin  and
               Weller, Orion  and
               Hallstr{\"o}m, Oskar  and
               Taghadouini, Said  and
               Gallagher, Alexis  and
               Biswas, Raja  and
               Ladhak, Faisal  and
               Aarsen, Tom  and
               Adams, Griffin Thomas  and
               Howard, Jeremy  and
               Poli, Iacopo},
  booktitle = {Proceedings of the 63rd Annual Meeting of the Association for Computational Linguistics (Volume 1: Long Papers)},
  month     = jul,
  year      = {2025},
  address   = {Vienna, Austria},
  publisher = {Association for Computational Linguistics},
  url       = {https://aclanthology.org/2025.acl-long.127/},
  doi       = {10.18653/v1/2025.acl-long.127},
  pages     = {2526--2547},
  isbn      = {979-8-89176-251-0}
}

@inproceedings{puri2021codenet,
  title     = {CodeNet: A Large-Scale {AI} for Code Dataset for Learning a Diversity of Coding Tasks},
  author    = {Ruchir Puri and David S Kung and Geert Janssen and Wei Zhang and Giacomo Domeniconi and Vladimir Zolotov and Julian Dolby and Jie Chen and Mihir Choudhury and Lindsey Decker and Veronika Thost and Luca Buratti and Saurabh Pujar and Shyam Ramji and Ulrich Finkler and Susan Malaika and Frederick Reiss},
  booktitle = {Thirty-fifth Conference on Neural Information Processing Systems Datasets and Benchmarks Track (Round 2)},
  year      = {2021},
  url       = {https://openreview.net/forum?id=6vZVBkCDrHT}
}

@misc{husain2020codesearchnet,
  title         = {CodeSearchNet Challenge: Evaluating the State of Semantic Code Search},
  author        = {Hamel Husain and Ho-Hsiang Wu and Tiferet Gazit and Miltiadis Allamanis and Marc Brockschmidt},
  year          = {2020},
  eprint        = {1909.09436},
  archiveprefix = {arXiv},
  primaryclass  = {cs.LG},
  url           = {https://arxiv.org/abs/1909.09436}
}

@inproceedings{wang2023codet5p,
  title     = {CodeT5+: Open Code Large Language Models for Code Understanding and Generation},
  author    = {Wang, Yue and Le, Hung and Gotmare, Akhilesh and Bui, Nghi and Li, Junnan and Hoi, Steven},
  booktitle = {Proceedings of the 2023 Conference on Empirical Methods in Natural Language Processing},
  pages     = {1069--1088},
  year      = {2023}
}

@misc{rozière2023codellama,
  title         = {Code Llama: Open Foundation Models for Code},
  author        = {Baptiste Rozière and Jonas Gehring and Fabian Gloeckle and Sten Sootla and Itai Gat and Xiaoqing Ellen Tan and Yossi Adi and Jingyu Liu and Romain Sauvestre and Tal Remez and Jérémy Rapin and Artyom Kozhevnikov and Ivan Evtimov and Joanna Bitton and Manish Bhatt and Cristian Canton Ferrer and Aaron Grattafiori and Wenhan Xiong and Alexandre Défossez and Jade Copet and Faisal Azhar and Hugo Touvron and Louis Martin and Nicolas Usunier and Thomas Scialom and Gabriel Synnaeve},
  year          = {2024},
  eprint        = {2308.12950},
  archiveprefix = {arXiv},
  primaryclass  = {cs.CL},
  url           = {https://arxiv.org/abs/2308.12950}
}

@inproceedings{feng-etal-2020-codebert,
  title     = {{C}ode{BERT}: A Pre-Trained Model for Programming and Natural Languages},
  author    = {Feng, Zhangyin  and
               Guo, Daya  and
               Tang, Duyu  and
               Duan, Nan  and
               Feng, Xiaocheng  and
               Gong, Ming  and
               Shou, Linjun  and
               Qin, Bing  and
               Liu, Ting  and
               Jiang, Daxin  and
               Zhou, Ming},
  booktitle = {Findings of the Association for Computational Linguistics: EMNLP 2020},
  year      = {2020},
  doi       = {10.18653/v1/2020.findings-emnlp.139},
  url       = {https://doi.org/10.18653/v1/2020.findings-emnlp.139}
}

@inproceedings{guo2022unixcoder,
  title     = {{U}ni{X}coder: Unified Cross-Modal Pre-training for Code Representation},
  author    = {Guo, Daya  and
               Lu, Shuai  and
               Duan, Nan  and
               Wang, Yanlin  and
               Zhou, Ming  and
               Yin, Jian},
  booktitle = {Proceedings of the 60th Annual Meeting of the Association for Computational Linguistics (Volume 1: Long Papers)},
  month     = may,
  year      = {2022},
  address   = {Dublin, Ireland},
  publisher = {Association for Computational Linguistics},
  url       = {https://aclanthology.org/2022.acl-long.499/},
  doi       = {10.18653/v1/2022.acl-long.499},
  pages     = {7212--7225}
}

@inproceedings{lyu2024PromptFuzz,
  author    = {Lyu, Yunlong and Xie, Yuxuan and Chen, Peng and Chen, Hao},
  title     = {Prompt Fuzzing for Fuzz Driver Generation},
  booktitle = {Proceedings of the 2024 on ACM SIGSAC Conference on Computer and Communications Security},
  year      = {2024},
  date      = {2024-10-14/2024-10-18},
  address   = {Salt Lake City, UT, USA},
  doi       = {10.1145/3658644.3670396},
  url       = {https://doi.org/10.1145/3658644.3670396}
}

@inproceedings{zhao2023understanding,
  title     = {Understanding Programs by Exploiting (Fuzzing) Test Cases},
  author    = {Zhao, Jianyu  and
               Rong, Yuyang  and
               Guo, Yiwen  and
               He, Yifeng  and
               Chen, Hao},
  booktitle = {Findings of the Association for Computational Linguistics: ACL 2023},
  month     = jul,
  year      = {2023},
  address   = {Toronto, Canada},
  publisher = {Association for Computational Linguistics},
  url       = {https://aclanthology.org/2023.findings-acl.678/},
  doi       = {10.18653/v1/2023.findings-acl.678},
  pages     = {10667--10679}
}

@article{kirkpatrick2017overcoming,
  author  = {James Kirkpatrick  and Razvan Pascanu  and Neil Rabinowitz  and Joel Veness  and Guillaume Desjardins  and Andrei A. Rusu  and Kieran Milan  and John Quan  and Tiago Ramalho  and Agnieszka Grabska-Barwinska  and Demis Hassabis  and Claudia Clopath  and Dharshan Kumaran  and Raia Hadsell },
  title   = {Overcoming catastrophic forgetting in neural networks},
  journal = {Proceedings of the National Academy of Sciences},
  volume  = {114},
  number  = {13},
  pages   = {3521-3526},
  year    = {2017},
  doi     = {10.1073/pnas.1611835114},
  url     = {https://www.pnas.org/doi/abs/10.1073/pnas.1611835114},
  eprint  = {https://www.pnas.org/doi/pdf/10.1073/pnas.1611835114}
}

@article{yu2022data,
  title    = {Data Augmentation by Program Transformation},
  journal  = {Journal of Systems and Software},
  volume   = {190},
  pages    = {111304},
  year     = {2022},
  issn     = {0164-1212},
  doi      = {10.1016/j.jss.2022.111304},
  url      = {https://doi.org/10.1016/j.jss.2022.111304},
  author   = {Shiwen Yu and Ting Wang and Ji Wang}
}

@inproceedings{devlin2019bert,
  title     = {{BERT}: Pre-training of Deep Bidirectional Transformers for Language Understanding},
  author    = {Devlin, Jacob  and
               Chang, Ming-Wei  and
               Lee, Kenton  and
               Toutanova, Kristina},
  booktitle = {Proceedings of the 2019 Conference of the North {A}merican Chapter of the Association for Computational Linguistics: Human Language Technologies, Volume 1 (Long and Short Papers)},
  month     = jun,
  year      = {2019},
  address   = {Minneapolis, Minnesota},
  publisher = {Association for Computational Linguistics},
  url       = {https://aclanthology.org/N19-1423/},
  doi       = {10.18653/v1/N19-1423},
  pages     = {4171--4186}
}

@misc{chen2021evaluating,
  title         = {Evaluating Large Language Models Trained on Code},
  author        = {Chen, Mark and Tworek, Jerry and Jun, Heewoo and Yuan, Qiming and Pinto, Henrique Ponde De Oliveira and Kaplan, Jared and Edwards, Harri and Burda, Yuri and Joseph, Nicholas and Brockman, Greg and others},
  year          = {2021},
  eprint        = {2107.03374},
  archiveprefix = {arXiv},
  primaryclass  = {cs.LG},
  url           = {https://arxiv.org/abs/2107.03374}
}

@inproceedings{he2024unitsyn,
  author    = {He, Yifeng and Huang, Jiabo and Rong, Yuyang and Guo, Yiwen and Wang, Ethan and Chen, Hao},
  title     = {UniTSyn: A Large-Scale Dataset Capable of Enhancing the Prowess of Large Language Models for Program Testing},
  year      = {2024},
  isbn      = {9798400706127},
  publisher = {Association for Computing Machinery},
  address   = {New York, NY, USA},
  url       = {https://doi.org/10.1145/3650212.3680342},
  doi       = {10.1145/3650212.3680342},
  booktitle = {Proceedings of the 33rd ACM SIGSOFT International Symposium on Software Testing and Analysis},
  pages     = {1061-1072},
  numpages  = {12},
  location  = {Vienna, Austria},
  series    = {ISSTA 2024}
}

@inproceedings{Ahmed2022multilingual,
  series     = {ICSE '22},
  title      = {Multilingual training for software engineering},
  url        = {http://dx.doi.org/10.1145/3510003.3510049},
  doi        = {10.1145/3510003.3510049},
  booktitle  = {Proceedings of the 44th International Conference on Software Engineering},
  publisher  = {ACM},
  author     = {Ahmed, Toufique and Devanbu, Premkumar},
  year       = {2022},
  month      = may,
  collection = {ICSE '22}
}

@inproceedings{yang2022natural_attack,
  author    = {Yang, Zhou and Shi, Jieke and He, Junda and Lo, David},
  title     = {Natural attack for pre-trained models of code},
  year      = {2022},
  isbn      = {9781450392211},
  publisher = {Association for Computing Machinery},
  address   = {New York, NY, USA},
  url       = {https://doi.org/10.1145/3510003.3510146},
  doi       = {10.1145/3510003.3510146},
  booktitle = {Proceedings of the 44th International Conference on Software Engineering},
  pages     = {1482-1493},
  numpages  = {12},
  location  = {Pittsburgh, Pennsylvania},
  series    = {ICSE '22}
}

@inproceedings{liu2023contrabert,
  title        = {{ContraBERT}: Enhancing Code Pre-trained Models via Contrastive Learning},
  author       = {Liu, Shangqing and Wu, Bozhi and Xie, Xiaofei and Meng, Guozhu and Liu, Yang},
  booktitle    = {2023 IEEE/ACM 45th International Conference on Software Engineering (ICSE)},
  pages        = {2476--2487},
  year         = {2023},
  organization = {IEEE},
  doi          = {10.1109/ICSE48619.2023.00207},
  url          = {https://doi.org/10.1109/ICSE48619.2023.00207}
}

@inproceedings{huang2024code,
  title     = {Code Representation Pre-training with Complements from Program Executions},
  author    = {Huang, Jiabo  and
               Zhao, Jianyu  and
               Rong, Yuyang  and
               Guo, Yiwen  and
               He, Yifeng  and
               Chen, Hao},
  booktitle = {Proceedings of the 2024 Conference on Empirical Methods in Natural Language Processing: Industry Track},
  month     = nov,
  year      = {2024},
  address   = {Miami, Florida, US},
  publisher = {Association for Computational Linguistics},
  url       = {https://aclanthology.org/2024.emnlp-industry.21/},
  doi       = {10.18653/v1/2024.emnlp-industry.21},
  pages     = {267--278}
}

@inproceedings{lu2021codexglue,
  title     = {Code{XGLUE}: A Machine Learning Benchmark Dataset for Code Understanding and Generation},
  author    = {Shuai Lu and Daya Guo and Shuo Ren and Junjie Huang and Alexey Svyatkovskiy and Ambrosio Blanco and Colin Clement and Dawn Drain and Daxin Jiang and Duyu Tang and Ge Li and Lidong Zhou and Linjun Shou and Long Zhou and Michele Tufano and MING GONG and Ming Zhou and Nan Duan and Neel Sundaresan and Shao Kun Deng and Shengyu Fu and Shujie LIU},
  booktitle = {Thirty-fifth Conference on Neural Information Processing Systems Datasets and Benchmarks Track (Round 1)},
  year      = {2021},
  url       = {https://openreview.net/forum?id=6lE4dQXaUcb}
}

@misc{he2025fuzzaug,
  title     = {{F}uzz{A}ug: Data Augmentation by Coverage-guided Fuzzing for Neural Test Generation},
  author    = {He, Yifeng  and
               Wang, Jicheng  and
               Rong, Yuyang  and
               Chen, Hao},
  editor    = {Christodoulopoulos, Christos  and
               Chakraborty, Tanmoy  and
               Rose, Carolyn  and
               Peng, Violet},
  booktitle = {Findings of the Association for Computational Linguistics: EMNLP 2025},
  month     = nov,
  year      = {2025},
  address   = {Suzhou, China},
  publisher = {Association for Computational Linguistics},
  url       = {https://aclanthology.org/2025.findings-emnlp.847/},
  doi       = {10.18653/v1/2025.findings-emnlp.847},
  pages     = {15642--15655},
  isbn      = {979-8-89176-335-7}
}

@inproceedings{bielik2020adversarial,
  title     = {Adversarial Robustness for Code},
  author    = {Bielik, Pavol and Vechev, Martin},
  booktitle = {Proceedings of the 37th International Conference on Machine Learning},
  pages     = {896--907},
  year      = {2020},
  editor    = {III, Hal Daumé and Singh, Aarti},
  volume    = {119},
  series    = {Proceedings of Machine Learning Research},
  month     = {13--18 Jul},
  publisher = {PMLR},
  url       = {https://proceedings.mlr.press/v119/bielik20a.html}
}

@inproceedings{guo2021graphcodebert,
  title     = {GraphCode{\{}BERT{\}}: Pre-training Code Representations with Data Flow},
  author    = {Daya Guo and Shuo Ren and Shuai Lu and Zhangyin Feng and Duyu Tang and Shujie LIU and Long Zhou and Nan Duan and Alexey Svyatkovskiy and Shengyu Fu and Michele Tufano and Shao Kun Deng and Colin Clement and Dawn Drain and Neel Sundaresan and Jian Yin and Daxin Jiang and Ming Zhou},
  booktitle = {International Conference on Learning Representations},
  year      = {2021},
  url       = {https://openreview.net/forum?id=jLoC4ez43PZ}
}

@inproceedings{wang2022bridging,
  author    = {Wang, Deze and Jia, Zhouyang and Li, Shanshan and Yu, Yue and Xiong, Yun and Dong, Wei and Liao, Xiangke},
  title     = {Bridging pre-trained models and downstream tasks for source code understanding},
  year      = {2022},
  isbn      = {9781450392211},
  publisher = {Association for Computing Machinery},
  address   = {New York, NY, USA},
  url       = {https://doi.org/10.1145/3510003.3510062},
  doi       = {10.1145/3510003.3510062},
  booktitle = {Proceedings of the 44th International Conference on Software Engineering},
  pages     = {287–298},
  numpages  = {12},
  location  = {Pittsburgh, Pennsylvania},
  series    = {ICSE '22}
}

@inproceedings{he2020momentum,
  title     = {Momentum contrast for unsupervised visual representation learning},
  author    = {He, Kaiming and Fan, Haoqi and Wu, Yuxin and Xie, Saining and Girshick, Ross},
  booktitle = {Proceedings of the IEEE/CVF conference on computer vision and pattern recognition},
  pages     = {9729--9738},
  year      = {2020},
  doi       = {10.1109/CVPR42600.2020.00975},
  url       = {https://doi.org/10.1109/CVPR42600.2020.00975}
}

@inproceedings{mou2016convolutional,
  author    = {Mou, Lili and Li, Ge and Zhang, Lu and Wang, Tao and Jin, Zhi},
  title     = {Convolutional neural networks over tree structures for programming language processing},
  year      = {2016},
  publisher = {AAAI Press},
  booktitle = {Proceedings of the Thirtieth AAAI Conference on Artificial Intelligence},
  pages     = {1287–1293},
  numpages  = {7},
  location  = {Phoenix, Arizona},
  series    = {AAAI'16},
  doi       = {10.1609/aaai.v30i1.10139},
  url       = {https://doi.org/10.1609/aaai.v30i1.10139}
}

@inproceedings{zhou2019devign,
  author    = {Zhou, Yaqin and Liu, Shangqing and Siow, Jingkai and Du, Xiaoning and Liu, Yang},
  booktitle = {Advances in Neural Information Processing Systems},
  publisher = {Curran Associates, Inc.},
  title     = {Devign: Effective Vulnerability Identification by Learning Comprehensive Program Semantics via Graph Neural Networks},
  url       = {https://proceedings.neurips.cc/paper/2019/hash/49265d2447bc3bbfe9e76306ce40a31f-Abstract.html},
  volume    = {32},
  year      = {2019}
}

@inproceedings{loshchilov2019decoupledweightdecayregularization,
  title     = {Decoupled Weight Decay Regularization},
  author    = {Ilya Loshchilov and Frank Hutter},
  year      = {2019},
  booktitle = {International Conference on Learning Representations},
  url       = {https://openreview.net/forum?id=Bkg6RiCqY7}
}

@inproceedings{musgrave2020metric,
  author    = {Musgrave, Kevin
               and Belongie, Serge
               and Lim, Ser-Nam},
  title     = {A Metric Learning Reality Check},
  booktitle = {European Conference on Computer Vision},
  year      = {2020},
  publisher = {Springer International Publishing},
  address   = {Cham},
  pages     = {681--699},
  isbn      = {978-3-030-58595-2},
  doi       = {10.1007/978-3-030-58595-2_41},
  url       = {https://link.springer.com/chapter/10.1007/978-3-030-58595-2_41}
}

@inproceedings{henkel2022semantic,
  author    = {Henkel, Jordan and Ramakrishnan, Goutham and Wang, Zi and Albarghouthi, Aws and Jha, Somesh and Reps, Thomas},
  booktitle = {2022 IEEE International Conference on Software Analysis, Evolution and Reengineering (SANER)},
  title     = {Semantic Robustness of Models of Source Code},
  year      = {2022},
  volume    = {},
  number    = {},
  pages     = {526-537},
  doi       = {10.1109/SANER53432.2022.00070}
}

@inproceedings{carlini2019secret,
  author    = {Carlini, Nicholas and Liu, Chang and Erlingsson, \'{U}lfar and Kos, Jernej and Song, Dawn},
  title     = {The Secret Sharer: Evaluating and Testing Unintended Memorization in Neural Networks},
  year      = {2019},
  isbn      = {9781939133069},
  publisher = {USENIX Association},
  address   = {USA},
  booktitle = {28th USENIX Security Symposium (USENIX Security 19)},
  pages     = {267-284},
  numpages  = {18},
  location  = {Santa Clara, CA, USA},
  series    = {SEC'19},
  url       = {https://www.usenix.org/conference/usenixsecurity19/presentation/carlini}
}

@inproceedings{ogura2018bring,
  author    = {Ogura, Naoto and Matsumoto, Shinsuke and Hata, Hideaki and Kusumoto, Shinji},
  booktitle = {2018 IEEE 25th International Conference on Software Analysis, Evolution and Reengineering (SANER)},
  title     = {Bring your own coding style},
  year      = {2018},
  volume    = {},
  number    = {},
  pages     = {527-531},
  doi       = {10.1109/SANER.2018.8330253}
}

@inproceedings{gao2021simcse,
  title     = {{S}im{CSE}: Simple Contrastive Learning of Sentence Embeddings},
  author    = {Gao, Tianyu  and
               Yao, Xingcheng  and
               Chen, Danqi},
  booktitle = {Proceedings of the 2021 Conference on Empirical Methods in Natural Language Processing},
  month     = nov,
  year      = {2021},
  address   = {Online and Punta Cana, Dominican Republic},
  publisher = {Association for Computational Linguistics},
  url       = {https://aclanthology.org/2021.emnlp-main.552/},
  doi       = {10.18653/v1/2021.emnlp-main.552},
  pages     = {6894--6910}
}

@misc{oord2019representation,
  title         = {Representation Learning with Contrastive Predictive Coding},
  author        = {Aaron van den Oord and Yazhe Li and Oriol Vinyals},
  year          = {2019},
  eprint        = {1807.03748},
  archiveprefix = {arXiv},
  primaryclass  = {cs.LG},
  url           = {https://arxiv.org/abs/1807.03748}
}

@inproceedings{kalra2024why,
  title     = {Why Warmup the Learning Rate? Underlying Mechanisms and Improvements},
  author    = {Dayal Singh Kalra and Maissam Barkeshli},
  booktitle = {The Thirty-eighth Annual Conference on Neural Information Processing Systems},
  year      = {2024},
  url       = {https://openreview.net/forum?id=NVl4SAmz5c}
}

@inproceedings{svajlenko2014towards,
  title        = {Towards a big data curated benchmark of inter-project code clones},
  author       = {Svajlenko, Jeffrey and Islam, Judith F and Keivanloo, Iman and Roy, Chanchal K and Mia, Mohammad Mamun},
  booktitle    = {2014 IEEE International Conference on Software Maintenance and Evolution},
  pages        = {476--480},
  year         = {2014},
  organization = {IEEE},
  doi          = {10.1109/ICSME.2014.77},
  url          = {https://doi.org/10.1109/ICSME.2014.77}
}

@inproceedings{krinke2022bigclonebenchharmful,
  author    = {Krinke, Jens and Ragkhitwetsagul, Chaiyong},
  booktitle = {2022 IEEE 16th International Workshop on Software Clones (IWSC)},
  title     = {BigCloneBench Considered Harmful for Machine Learning},
  year      = {2022},
  volume    = {},
  number    = {},
  pages     = {1-7},
  doi       = {10.1109/IWSC55060.2022.00008}
}

@misc{krinke2025misuse,
  title         = {How the Misuse of a Dataset Harmed Semantic Clone Detection},
  author        = {Jens Krinke and Chaiyong Ragkhitwetsagul},
  year          = {2025},
  eprint        = {2505.04311},
  archiveprefix = {arXiv},
  primaryclass  = {cs.SE},
  url           = {https://arxiv.org/abs/2505.04311}
}

@inproceedings{pmlr-v119-chen20j,
  title     = {A Simple Framework for Contrastive Learning of Visual Representations},
  author    = {Chen, Ting and Kornblith, Simon and Norouzi, Mohammad and Hinton, Geoffrey},
  booktitle = {Proceedings of the 37th International Conference on Machine Learning},
  pages     = {1597--1607},
  year      = {2020},
  editor    = {III, Hal Daumé and Singh, Aarti},
  volume    = {119},
  series    = {Proceedings of Machine Learning Research},
  month     = {13--18 Jul},
  publisher = {PMLR},
  url       = {https://proceedings.mlr.press/v119/chen20j.html}
}

@inproceedings{chen2022varclr,
  author    = {Chen, Qibin and Lacomis, Jeremy and Schwartz, Edward J. and Neubig, Graham and Vasilescu, Bogdan and Le Goues, Claire},
  title     = {{V}ar{CLR}: variable semantic representation pre-training via contrastive learning},
  year      = {2022},
  isbn      = {9781450392211},
  publisher = {Association for Computing Machinery},
  address   = {New York, NY, USA},
  url       = {https://doi.org/10.1145/3510003.3510162},
  doi       = {10.1145/3510003.3510162},
  booktitle = {Proceedings of the 44th International Conference on Software Engineering},
  pages     = {2327–2339},
  numpages  = {13},
  location  = {Pittsburgh, Pennsylvania},
  series    = {ICSE '22}
}

@inproceedings{dui2021selfsupervised,
  author    = {Bui, Nghi D. Q. and Yu, Yijun and Jiang, Lingxiao},
  title     = {Self-Supervised Contrastive Learning for Code Retrieval and Summarization via Semantic-Preserving Transformations},
  year      = {2021},
  isbn      = {9781450380379},
  publisher = {Association for Computing Machinery},
  address   = {New York, NY, USA},
  url       = {https://doi.org/10.1145/3404835.3462840},
  doi       = {10.1145/3404835.3462840},
  booktitle = {Proceedings of the 44th International ACM SIGIR Conference on Research and Development in Information Retrieval},
  pages     = {511–521},
  numpages  = {11},
  location  = {Virtual Event, Canada},
  series    = {SIGIR '21}
}

@inproceedings{jain2021contrastive,
  title     = {Contrastive Code Representation Learning},
  author    = {Jain, Paras  and
               Jain, Ajay  and
               Zhang, Tianjun  and
               Abbeel, Pieter  and
               Gonzalez, Joseph  and
               Stoica, Ion},
  booktitle = {Proceedings of the 2021 Conference on Empirical Methods in Natural Language Processing},
  month     = nov,
  year      = {2021},
  address   = {Online and Punta Cana, Dominican Republic},
  publisher = {Association for Computational Linguistics},
  url       = {https://aclanthology.org/2021.emnlp-main.482/},
  doi       = {10.18653/v1/2021.emnlp-main.482},
  pages     = {5954--5971}
}

@article{yefet2020damp,
  author     = {Yefet, Noam and Alon, Uri and Yahav, Eran},
  title      = {Adversarial examples for models of code},
  year       = {2020},
  issue_date = {November 2020},
  publisher  = {Association for Computing Machinery},
  address    = {New York, NY, USA},
  volume     = {4},
  number     = {OOPSLA},
  url        = {https://doi.org/10.1145/3428230},
  doi        = {10.1145/3428230},
  journal    = {Proc. ACM Program. Lang.},
  month      = nov,
  articleno  = {162},
  numpages   = {30}
}

@article{zhang2020generating,
  title   = {Generating Adversarial Examples for Holding Robustness of Source Code Processing Models},
  volume  = {34},
  url     = {https://ojs.aaai.org/index.php/AAAI/article/view/5469},
  doi     = {10.1609/aaai.v34i01.5469},
  number  = {01},
  journal = {Proceedings of the AAAI Conference on Artificial Intelligence},
  author  = {Zhang, Huangzhao and Li, Zhuo and Li, Ge and Ma, Lei and Liu, Yang and Jin, Zhi},
  year    = {2020},
  month   = {Apr.},
  pages   = {1169-1176}
}

@article{tSNE,
  author  = {Laurens van der Maaten and Geoffrey Hinton},
  title   = {Visualizing Data using t-SNE},
  journal = {Journal of Machine Learning Research},
  year    = {2008},
  volume  = {9},
  number  = {86},
  pages   = {2579--2605},
  url     = {http://jmlr.org/papers/v9/vandermaaten08a.html}
}

@inproceedings{khosla2020supervised,
  author    = {Khosla, Prannay and Teterwak, Piotr and Wang, Chen and Sarna, Aaron and Tian, Yonglong and Isola, Phillip and Maschinot, Aaron and Liu, Ce and Krishnan, Dilip},
  title     = {Supervised contrastive learning},
  year      = {2020},
  isbn      = {9781713829546},
  publisher = {Curran Associates Inc.},
  address   = {Red Hook, NY, USA},
  booktitle = {Proceedings of the 34th International Conference on Neural Information Processing Systems},
  articleno = {1567},
  numpages  = {13},
  location  = {Vancouver, BC, Canada},
  series    = {NIPS '20}
}

@inproceedings{ding2024primevul,
  author    = {Ding, Yangruibo and Fu, Yanjun and Ibrahim, Omniyyah and Sitawarin, Chawin and Chen, Xinyun and Alomair, Basel and Wagner, David and Ray, Baishakhi and Chen, Yizheng},
  title     = {Vulnerability Detection with Code Language Models: How Far are We?},
  booktitle = {2025 IEEE/ACM 47th International Conference on Software Engineering (ICSE)},
  series    = {ICSE},
  year      = {2025},
  doi       = {10.1109/ICSE55347.2025.00038},
  url       = {https://doi.org/10.1109/ICSE55347.2025.00038}
}

@article{thennal2025overparameterized,
  title     = {Large Language Models Are Overparameterized Text Encoders},
  author    = {K, Thennal D  and
               Fischer, Tim  and
               Biemann, Chris},
  booktitle = {Proceedings of the 10th Workshop on Representation Learning for NLP (RepL4NLP-2025)},
  month     = may,
  year      = {2025},
  address   = {Albuquerque, NM},
  publisher = {Association for Computational Linguistics},
  url       = {https://aclanthology.org/2025.repl4nlp-1.13/},
  doi       = {10.18653/v1/2025.repl4nlp-1.13},
  pages     = {170--184},
  isbn      = {979-8-89176-245-9}
}

@article{lei2025efficient,
  title   = {Making Large Language Models Efficient Dense Retrievers},
  author  = {Lei, Yibin and He, Shwai and Li, Ang and Yates, Andrew},
  journal = {arXiv preprint arXiv:2512.20612},
  year    = {2025},
  url     = {https://arxiv.org/abs/2512.20612}
}

@inproceedings{chakraborty2022natgen,
  author    = {Chakraborty, Saikat and Ahmed, Toufique and Ding, Yangruibo and Devanbu, Premkumar T. and Ray, Baishakhi},
  title     = {{NatGen}: generative pre-training by "naturalizing" source code},
  year      = {2022},
  isbn      = {9781450394130},
  publisher = {Association for Computing Machinery},
  address   = {New York, NY, USA},
  url       = {https://doi.org/10.1145/3540250.3549162},
  doi       = {https://doi.org/10.1145/3540250.3549162},
  booktitle = {Proceedings of the 30th ACM Joint European Software Engineering Conference and Symposium on the Foundations of Software Engineering},
  pages     = {18–30},
  numpages  = {13},
  location  = {Singapore, Singapore},
  series    = {ESEC/FSE 2022}
}

@article{rabin2021generalizability,
  title    = {On the generalizability of Neural Program Models with respect to semantic-preserving program transformations},
  journal  = {Information and Software Technology},
  volume   = {135},
  pages    = {106552},
  year     = {2021},
  issn     = {0950-5849},
  doi      = {https://doi.org/10.1016/j.infsof.2021.106552},
  url      = {https://www.sciencedirect.com/science/article/pii/S0950584921000379},
  author   = {Md Rafiqul Islam Rabin and Nghi D.Q. Bui and Ke Wang and Yijun Yu and Lingxiao Jiang and Mohammad Amin Alipour}
}

@article{lin2026cl4d,
  title        = {Towards Better Code Understanding in Decoder-Only Models with Contrastive Learning},
  volume       = {40},
  url          = {https://ojs.aaai.org/index.php/AAAI/article/view/40471},
  doi          = {10.1609/aaai.v40i38.40471},
  abstractnote = {Recent advances in large-scale code generation models have led to remarkable progress in producing high-quality code. These models are trained in a self-supervised manner on extensive unlabeled code corpora using a decoder-only architecture. However, despite their generative strength, decoder-only models often exhibit limited performance on code understanding tasks such as code search and clone detection, primarily due to their generation-oriented training objectives. While training large encoder-only models from scratch on massive code datasets can improve understanding ability but remains computationally expensive and time-consuming.
                  In this paper, we explore a more efficient alternative by transferring knowledge from pre-trained decoder-only code generation models to code understanding tasks. We investigate how decoder-only architectures can be effectively adapted to learn discriminative and semantically meaningful code representations. To this end, we propose CL4D, a contrastive learning framework tailored to strengthen the representation capabilities of decoder-only models.
                  Extensive experiments on multiple benchmark datasets demonstrate that CL4D achieves competitive or superior performance compared to existing methods on representative code understanding tasks, including code search and clone detection. Further analysis reveals that CL4D substantially improves the semantic alignment of code representations by reducing the distance between semantically similar code snippets. These findings highlight the feasibility of leveraging decoder-only models as a unified backbone for both code generation and understanding.},
  number       = {38},
  journal      = {Proceedings of the AAAI Conference on Artificial Intelligence},
  author       = {Lin, Jiayi and Wang, Yanlin and Yang, Yibiao and Zhang, Lei and Xie, Yutao},
  year         = {2026},
  month        = {Mar.},
  pages        = {32006-32014}
}

@misc{jina_code_embeddings_2025,
  author       = {{Jina AI}},
  title        = {Jina Code Embeddings},
  year         = {2025},
  howpublished = {\url{https://huggingface.co/jinaai/jina-embeddings-v2-base-code}},
  note         = {Code embedding model}
}

@misc{qwen3_embedding_2025,
  author       = {{Qwen Team}},
  title        = {Qwen3 Embedding},
  year         = {2025},
  howpublished = {\url{https://huggingface.co/Qwen}},
  note         = {Embedding model series}
}

@misc{embeddinggemma_2025,
  author       = {{Google}},
  title        = {EmbeddingGemma},
  year         = {2025},
  howpublished = {\url{https://huggingface.co/google}},
  note         = {Open embedding model}
}

@misc{coderankembed_2024,
  author       = {{Nomic AI}},
  title        = {CodeRankEmbed},
  year         = {2024},
  howpublished = {\url{https://huggingface.co/nomic-ai/CodeRankEmbed}},
  note         = {Code retrieval embedding model}
}
